\documentclass[10pt,twocolumn,letterpaper]{article}

\usepackage{cvpr}
\usepackage{times}
\usepackage{array}
\usepackage{epsfig}
\usepackage{graphicx}
\usepackage{amsmath}
\usepackage{amssymb}
\usepackage{setspace}
\usepackage{algorithm}
\usepackage{algorithmic}
\usepackage{multirow}
\graphicspath{
    {figure/}
}

\usepackage[breaklinks=true,bookmarks=false]{hyperref}

\cvprfinalcopy % *** Uncomment this line for the final submission

\def\cvprPaperID{****} % *** Enter the CVPR Paper ID here

\ifcvprfinal\fi
\begin{document}

%%%%%%%%% TITLE
\title{RON: Reverse Connection with Objectness Prior Networks \\for Object Detection}

\author{Tao Kong$^{1*}$ ~~  Fuchun Sun$^1$ ~~ Anbang Yao$^2$ ~~ Huaping Liu$^1$  ~~ Ming Lu$^3$~~ Yurong Chen$^2$\\
% For a paper whose aut lines up until the closing ``}''.
$^1$State Key Lab. of Intelligent Technology and Systems, Tsinghua National Laboratory for \\
Information Science and Technology (TNList), Department of Computer Science and Technology, \\
Tsinghua University $^2$Intel Labs China $^3$Department of Electronic Engineering, Tsinghua University\\
\tt\small $^{1}$\{kt14@mails,fcsun@,hpliu@\}.tsinghua.edu.cn \\
\tt\small $^2$\{anbang.yao, yurong.chen\}@.intel.com ~~ $^3$lu-m13@mails.tsinghua.edu.cn
}
\maketitle
%\thispagestyle{empty}

%%%%%%%%% ABSTRACT
\begin{abstract}
\let\thefootnote\relax\footnotetext{*This work was done when Tao Kong was an intern at Intel Labs China supervised by Anbang Yao who is responsible for correspondence.}
We present RON, an efficient and effective framework for generic object detection. Our motivation is to smartly associate the best of the region-based (e.g., Faster R-CNN) and region-free (e.g., SSD) methodologies. Under fully convolutional architecture, RON mainly focuses on two fundamental problems: (a) multi-scale object localization and (b) negative sample mining. To address (a), we design the reverse connection, which enables the network to detect objects on multi-levels of CNNs. To deal with (b), we propose the objectness prior to significantly reduce the searching space of objects. We optimize the reverse connection, objectness prior and object detector jointly by a multi-task loss function, thus RON can directly predict final detection results from all locations of various feature maps.

Extensive experiments on the challenging PASCAL VOC 2007, PASCAL VOC 2012 and MS COCO benchmarks demonstrate the competitive performance of RON. Specifically, with VGG-16 and low resolution 384$\times$384 input size, the network gets 81.3\% mAP on PASCAL VOC 2007, 80.7\% mAP on PASCAL VOC 2012 datasets. Its superiority increases when datasets become larger and more difficult, as demonstrated by the results on the MS COCO dataset. With 1.5G GPU memory at test phase, the speed of the network is 15 FPS, 3$\times$ faster than the Faster R-CNN counterpart. Code will be available at \url{https://github.com/taokong/RON}.

\end{abstract}

%%%%%%%%% BODY TEXT
\section{Introduction}

\begin{figure}[h]
\begin{center}
    \includegraphics[width=0.8\linewidth]{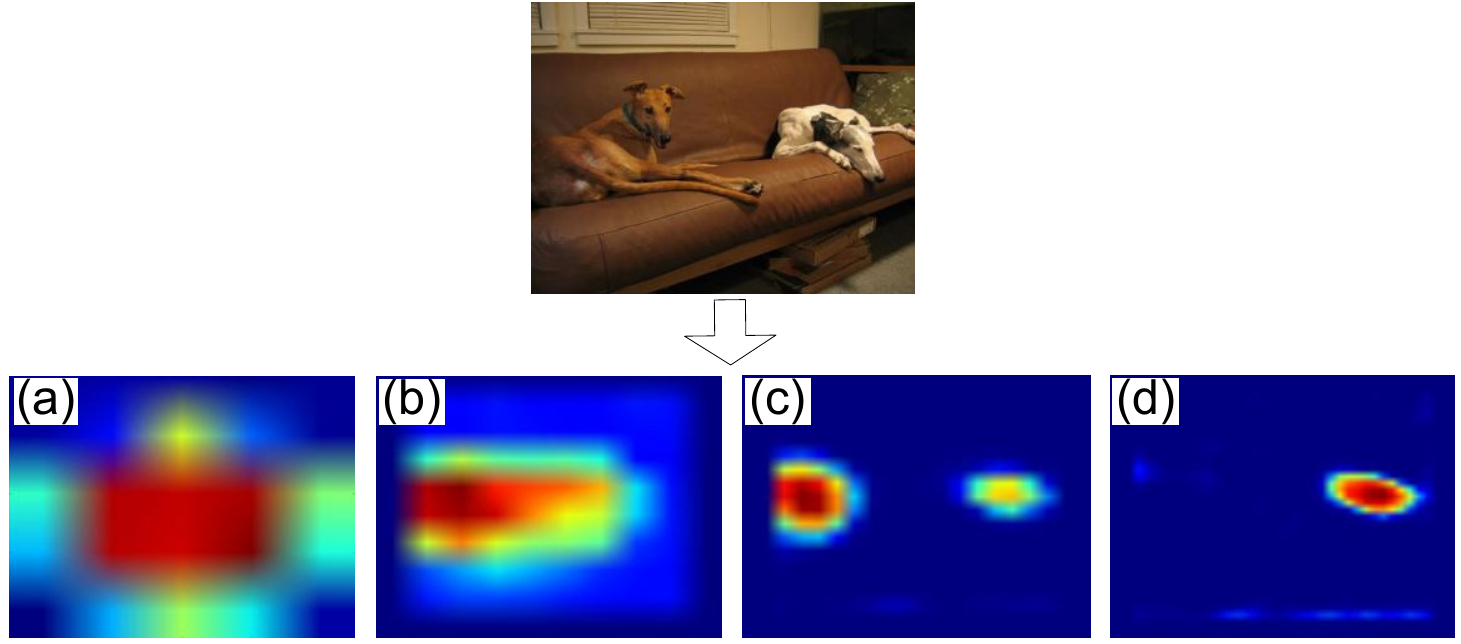}
\end{center}
\vskip -0.1 in
   \caption{Objectness prior generated from a specific image. In this example, sofa is responded at scales (a) and (b), the brown dog is responded at scale (c) and the white spotted dog is responded at scale (d). The network will generate detection results with the guidance of objectness prior .}
\label{pipline}
\vskip -0.2 in
\end{figure}

We are witnessing  significant advances in object detection area, mainly thanks to the deep networks. Current top deep-networks-based object detection frameworks could be grouped into two main streams: the region-based methods \cite{girshick2014rich}\cite{fasterrcnn}\cite{frcnn}\cite{hypernet} and the region-free methods \cite{yolo}\cite{ssd}.

The region-based methods divide the object detection task into two sub-problems: At the first stage, a dedicated region proposal generation network is grafted on deep convolutional neural networks (CNNs) which could generate high quality candidate boxes. Then at the second stage, a region-wise subnetwork is designed to classify and refine these candidate boxes. Using very deep CNNs \cite{resnet}\cite{vgg16}, the Fast R-CNN pipeline \cite{frcnn}\cite{fasterrcnn} has recently shown high accuracy on mainstream object detection benchmarks \cite{pascalvoc}\cite{imagenet}\cite{coco}. The region proposal stage could reject most of the background samples, thus the searching space for object detection is largely reduced \cite{mcg}\cite{edgeboxes}. Multi-stage training process is usually developed for joint optimization of region proposal generation and post detection (e.g., \cite{rfcn}\cite{fasterrcnn}\cite{hypernet}).  In Fast R-CNN \cite{frcnn}, the region-wise subnetwork repeatedly evaluates thousands of region proposals to generate detection scores. Under Fast R-CNN pipeline, Faster R-CNN shares full-image convolutional features with the detection network to enable nearly cost-free region proposals. Recently, R-FCN \cite{rfcn} tries to make the unshared per-RoI computation of Faster R-CNN to be sharable by adding position-sensitive score maps. Nevertheless, R-FCN still needs region proposals generated from region proposal networks \cite{fasterrcnn}. To ensure detection accuracy, all methods resize the image to a large enough size (usually with the shortest side of 600 pixels). It is somewhat resource/time consuming when feeding the image into deep networks, both in training  and inference time. For example, predicting with Faster R-CNN usually takes 0.2 s per image using about 5GB GPU memory for VGG-16 networks \cite{vgg16}.

Another solution family is the region-free methods \cite{yolo}\cite{ssd}. These methods treat object detection as a single shot problem, straight from image pixels to bounding box coordinates by fully convolutional networks (FCNs). The main advantage of these detectors is high efficiency. Originated from YOLO \cite{yolo}, SSD \cite{ssd} tries to deal object detection problem with multiple layers of deep CNNs. With low resolution input, the SSD detector could get state-of-the-art detection results. However, the detection accuracy of these methods still has room for improvement: (a) Without region proposals, the detectors have to suppress all of the negative candidate boxes only at the detection module. It will increase the difficulties on training the detection module. (b) YOLO detects objects with the top-most CNN layer, without deeply exploring the detection capacities of different layers. SSD tries to improve the detection performance by adding former layers' results. However, SSD still struggles with small instances, mainly because of the limited information of middle layers. These two main bottlenecks affect the detection accuracy of the methods.

Driven by the success of the two solution families, a critical question arises: \emph{is it possible to develop an elegant framework which can smartly associate the best of both methodologies and eliminate their major demerits?} We answer this question by trying to bridge the gap between the region-based and region-free methodologies. To achieve this goal, we focus on two fundamental problems: (a) Multi-scale object localization. Objects of various scales could appear at any position of an image, so tens of thousands of regions with different positions/scales/aspect ratios should be considered. Prior works \cite{hypernet}\cite{ion} show that multi-scale representation will significantly improve object detection of various scales. However, these methods always detect \emph{various scales of objects} at \emph{one layer} of a network \cite{hypernet}\cite{fasterrcnn}\cite{rfcn}. With the proposed reverse connection, objects are detected \emph{on their corresponding network scales}, which is more elegant and easier to optimize. (b) Negative space mining. The ratio between object and non-object samples is seriously imbalanced. So an object detector should have effective negative mining strategies \cite{ohem}. In order to reduce the searching space of the objects, we create an objectness prior (Figure \ref{pipline}) on convolutional feature maps and jointly optimize it with the detector at training phase.

 As a result, we propose RON (Reverse connection with Objectness prior Networks) object detection framework, which could associate the merits of region-based and region-free approaches. Furthermore, recent complementary advances such as hard example mining \cite{ohem}, bounding box regression \cite{girshick2014rich} and multi-layer representation \cite{hypernet}\cite{ion} could be naturally employed.

\textbf{Contributions}. We make the following contributions:
\begin{enumerate}
\item We propose RON, a fully convolutional framework for end-to-end object detection. Firstly, the reverse connection assists the former layers of CNNs with more semantic information. Second, the objectness prior gives an explicit guide to the searching of objects. Finally, the multi-task loss function enables us to optimize the whole network end-to-end on detection performance.
\item In order to achieve high detection accuracy, effective training strategies like negative example mining and data augmentation have been employed. With low resolution 384$\times$384 input size, RON achieves state-of-the-art results on PASCAL VOC 2007, with a 81.3\% mAP, VOC 2012, with a 80.7\% mAP, and MS COCO, with a 27.4\% mAP.
\item RON is time and resource efficient. With 1.5G GPU memory, the total feed-forward speed is 15 FPS, 3$\times$ faster than the seminal Faster R-CNN.  Moreover, we conduct extensive design choices like the layers combination, with/without objectness prior, and other variations.
\end{enumerate}

\section{Related Work}

Object detection is a fundamental and heavily-researched task in computer vision area. It aims to localize and recognize every object instance with a bounding box \cite{pascalvoc}\cite{coco}. Before the success of deep CNNs \cite{alexnet}, the widely used detection systems are based on the combination of independent components (HOG \cite{hog}, SIFT \cite{sift} et al.). The DPM \cite{dpm} and its variants \cite{dpm1}\cite{fastdpm} have been the dominant methods for years. These methods use object component descriptors as features and sweep through the entire image to find regions with a class-specific maximum response. With the great success of the deep learning on large scale object recognition, several  works based on CNNs have been proposed \cite{cnn1}\cite{overfeat}. R-CNN \cite{girshick2014rich} and its variants usually combine region proposals (generated from Selective Search \cite{ss}, Edgeboxes \cite{edgeboxes}, MCG \cite{mcg}, et al.) and ConvNet based post-classification. These methods have brought dramatic improvement on detection accuracy \cite{alexnet}. After the original R-CNN, researches are improving it in a variety of ways. The SPP-Net \cite{sppnet} and Fast R-CNN \cite{frcnn} speed up the R-CNN approach with RoI-Pooling (Spatial-Pyramid-Pooling) that allows the classification layers to reuse features computed over CNN feature maps. Under Fast R-CNN pipeline, several works try to improve the detection speed and accuracy, with more effective region proposals \cite{fasterrcnn}, multi-layer fusion \cite{ion}\cite{hypernet}, context information \cite{resnet}\cite{mrcnn} and more effective training strategy \cite{ohem}. R-FCN \cite{rfcn} tries to reduce the computation time with position-sensitive score maps under ResNet architecture \cite{resnet}. Thanks to the FCNs \cite{fcn} which could give position-wise information of input images, several works are trying to solve the object detection problem by FCNs. The methods skip the region proposal generation step and predict bounding boxes and detection confidences of multiple categories directly \cite{multibox}. YOLO \cite{yolo} uses the top most CNN feature maps to predict both confidences and locations for  multiple categories. Originated from YOLO, SSD \cite{ssd} tries to predict detection results at multiple CNN layers. With carefully designed training strategies, SSD could get competitive detection results. The main advantage of these methods is high time-efficiency. For example, the feed-forward speed of YOLO is 45 FPS, 9$\times$ faster than Faster R-CNN.

\section{Network Architecture}\label{network}

\begin{figure}[t]
\begin{center}
    \includegraphics[width=0.95\linewidth]{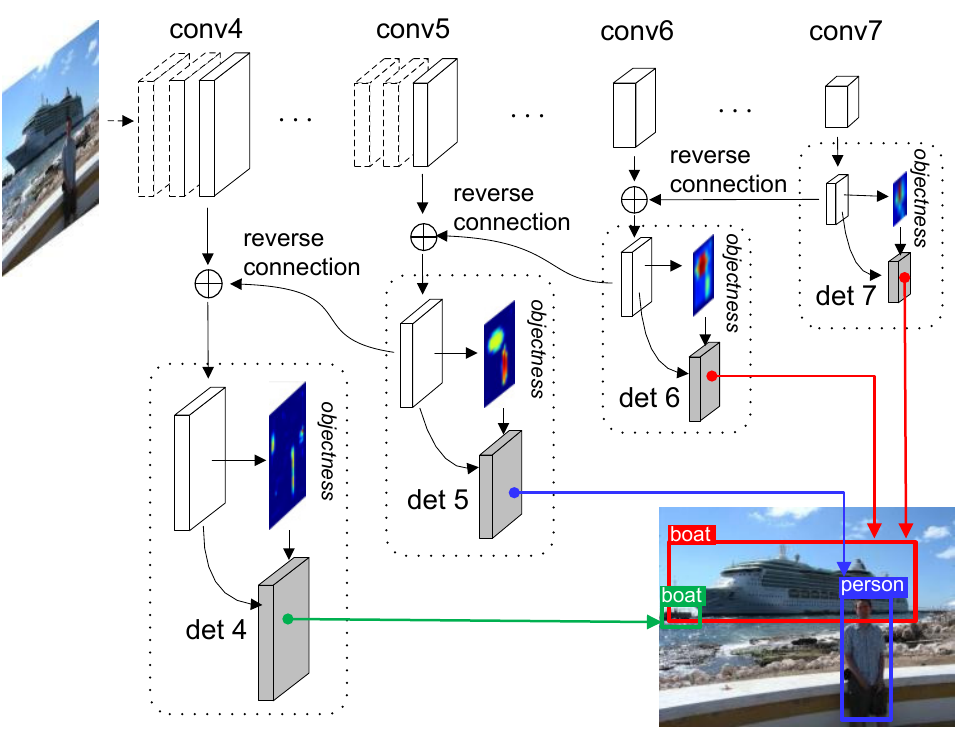}
\end{center}
\vskip -0.1 in
   \caption{RON object detection overview. Given an input image, the network firstly computes features of the backbone network. Then at each detection scale: (a) adds reverse connection; (b) generates objectness prior; (c) detects object on its corresponding CNN scales and locations. Finally, all detection results are fused and selected with non-maximum suppression.}
\label{framework}
\vskip -0.2 in
\end{figure}

This section describes RON object detection framework (Figure \ref{framework}). We first introduce the reverse connection on traditional CNNs in Section \ref{multiscale}, such that different network scales have effective detection capabilities. Then in Section \ref{refboxes}, we explain how to generate candidate boxes on different network scales. Next, we present the objectness prior to guide the search of objects in Section \ref{objprior} and Secction \ref{detection_sec}. Finally, we combine the objectness prior and object detection into a unified network for joint training and testing (Section \ref{combine}).

\textbf{Network preparation} We use VGG-16 as the test case reference model, which  is pre-trained with ImageNet dataset \cite{vgg16}. Recall that VGG-16 has 13 convolutional layers and 3 fully-connected layers. We convert FC6 (14th layer) and FC7 (15th layer) to convolutional layers \cite{fcn}, and use 2$\times$2 convolutional kernels with stride 2 to reduce the resolution of FC7 by half\footnote{The last FC layer (16th layer) is not used in this paper.}. By now, the feature map sizes used for object detection are $1/8$ (conv 4\_3), $1/16$ (conv 5\_3), $1/32$ (conv 6) and $1/64$ (conv 7) of the input size, both in width and height (see Figure \ref{framework} top).

\subsection{Reverse Connection}\label{multiscale}

Combining fine-grained details with highly-abstracted information helps object detection with different scales \cite{hypercol}\cite{fcn}\cite{resnet}. The region-based networks usually fuse multiple CNN layers into a single feature map \cite{ion}\cite{hypernet}. Then object detection is performed on the fused maps with region-wise subnetworks \cite{frcnn}. As all objects need to be detected based on the fixed features, the optimization becomes much complex. SSD \cite{ssd} detects objects on multiple CNN layers. However, the semantic information of former layers is limited, which affects the detection performance of these layers. This is the reason why SSD has much worse performance on smaller objects than bigger objects \cite{ssd}.

\begin{figure}
\begin{center}
    \includegraphics[width=0.5\linewidth]{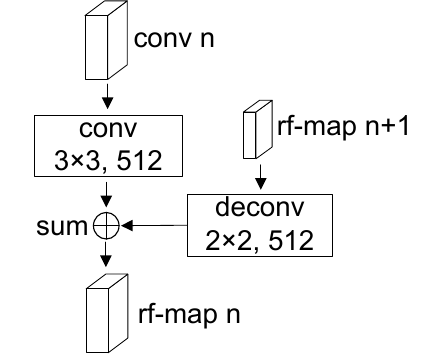}
\end{center}
\vskip -0.1 in
\caption{A reverse connection block.}
\label{block}
\vskip -0.2 in
\end{figure}

Inspired from the success of residual connection \cite{resnet} which eases the training of much deeper networks, we propose the reverse connection on traditional CNN architectures. The reverse connection enables former features to have more semantic information. One reverse connection block is shown in Figure \ref{block}. Firstly, a deconvolutional layer is applied to the reverse fusion map (annotated as rf-map) $n+1$, and a convolutional layer is grafted on backbone layer $n$ to guarantee the inputs have the same dimension. Then the two corresponding maps are merged by element-wise addition. The reverse fusion map 7 is the convolutional output (with 512 channels by 3$\times$3 kernels) of the backbone layer 7. After this layer has been generated, each reverse connection block will be generated in the same way, as shown in Figure \ref{framework}. In total, there are four reverse fusion maps with different scales.

Compared with methods using single layer for object detection \cite{hypernet}\cite{fasterrcnn}, multi-scale representation is more effective on locating all scales of objects (as shown in experiments). More importantly, as the reverse connection is learnable, the semantic information of former layers can be significantly enriched.  This characteristic makes RON more effective in detecting all scales of objects compared with \cite{ssd}.

\subsection{Reference Boxes}\label{refboxes}

In this section, we describe how to generate bounding boxes on feature maps produced from Section \ref{multiscale}. Feature maps from different levels within a network are known to have different receptive field sizes \cite{sppnet}. As the reverse connection can generate multiple feature maps with different scales. We can design the distribution of boxes so that specific feature map locations can be learned to be responsive to particular scales of objects. Denoting the minimum scale with $s_{min}$, the scales $S_k$ of boxes at each feature map $k$ are
\begin{equation}
S_k = \{(2k-1)\cdot s_{min},  2k\cdot s_{min}\}, k\in\{1,2,3,4\}.
\label{boxes}
\end{equation}
We also impose different aspect ratios \{$\frac{1}{3}, \frac{1}{2}, 1, 2, 3$\} for the default boxes. The width and height of each box are computed with respect to the aspect ratio \cite{fasterrcnn}. In total, there are 2 scales and 5 aspect ratios at each feature map location. The $s_{min}$ is $\frac{1}{10}$ of the input size (e.g., 32 pixels for 320$\times$320 model). By combining predictions for all default boxes with different scales and aspect ratios, we have a diverse set of predictions, covering various object sizes and shapes.

\subsection{Objectness Prior}\label{objprior}

As shown in Section \ref{refboxes}, we consider default boxes with different scales and aspect ratios from many feature maps. However, only a tiny fraction of boxes covers objects. In other words, the ratio between object and non-object samples is seriously imbalanced. The region-based methods overcome this problem by region proposal networks \cite{hypernet}\cite{fasterrcnn}. However, the region proposals will bring translation variance compared with the default boxes. So the Fast R-CNN pipeline usually uses region-wise networks for post detection, which brings repeated computation \cite{rfcn}. In contrast, we add an objectness prior for guiding the search of objects, without generating new region proposals. Concretely, we add a $3$$\times3\times$$2$ convolutional layer followed by a Softmax function to indicate the  existence of an object in each box. The channel number of objectness prior maps is 10, as there are 10 default boxes at each location.

Figure \ref{pipline} shows the multi-scale objectness prior generated from a specific image. For visualization, the objectness prior maps are averaged along the channel dimension. We see that the objectness prior maps could explicitly reflect the existence of an object. Thus the searching space of the objects could be dramatically reduced. Objects of various scales will respond at their corresponding feature maps, and we enable this by appropriate matching and end-to-end training. More results could be seen in the experiment section.

\subsection{Detection and Bounding Box Regression}\label{detection_sec}

\begin{figure}
\begin{center}
    \includegraphics[width=0.38\linewidth]{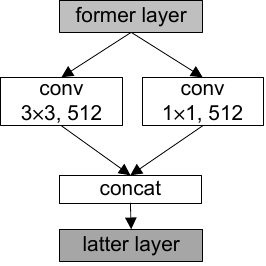}
\end{center}
\vskip -0.1 in
   \caption{One inception block.}
\label{incept}
\vskip -0.1 in
\end{figure}

Different from objectness prior, the detection module needs to classify regions into $K$$+1$ categories ($K$$=20$ for PASCAL VOC dataset and $K$$=80$ for MS COCO dataset, plus 1 for background). We employ the inception module \cite{googlenet} on the feature maps generated in Section \ref{multiscale} to perform detection. Concretely, we add two inception blocks (one block is shown in Figure \ref{incept}) on feature maps, and classify the final inception outputs. There are many inception choices as shown in \cite{googlenet}. In this paper, we just use the most simple structure.

\begin{figure}[h]
\begin{center}
    \includegraphics[width=0.8\linewidth]{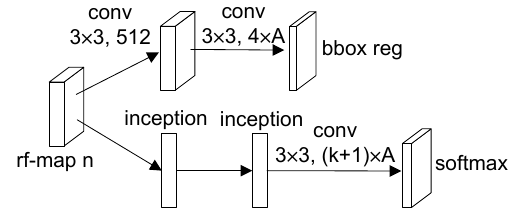}
\end{center}
\vskip -0.1 in
   \caption{Object detection and bounding box regression modules. Top: bounding box regression; Bottom: object classification.}
\label{det_module}
\vskip -0.2 in
\end{figure}

With Softmax, the sub-network outputs the per-class score that indicates the presence of a class-specific instance. For bounding box regression, we predict the offsets relative to the default boxes in the cell (see Figure \ref{det_module}).

\subsection{Combining Objectness Prior with Detection}\label{combine}

In this section, we explain how RON combines objectness prior with object detection. We assist object detection with objectness prior both in training and testing phases. For training the network, we firstly assign a binary class label to each candidate region generated from Section \ref{refboxes}. Then if the region covers object, we also assign a class-specific label to it. For each ground truth  box, we (i) match it with the candidate region with most jaccard overlap; (ii) match candidate regions to any ground truth with jaccard overlap higher than 0.5. This matching strategy guarantees that each ground truth box has at least one region box assigned to it. We assign negative labels to the boxes with jaccard overlap lower than 0.3.

\begin{figure}[h]
\begin{center}
    \includegraphics[width=0.6\linewidth]{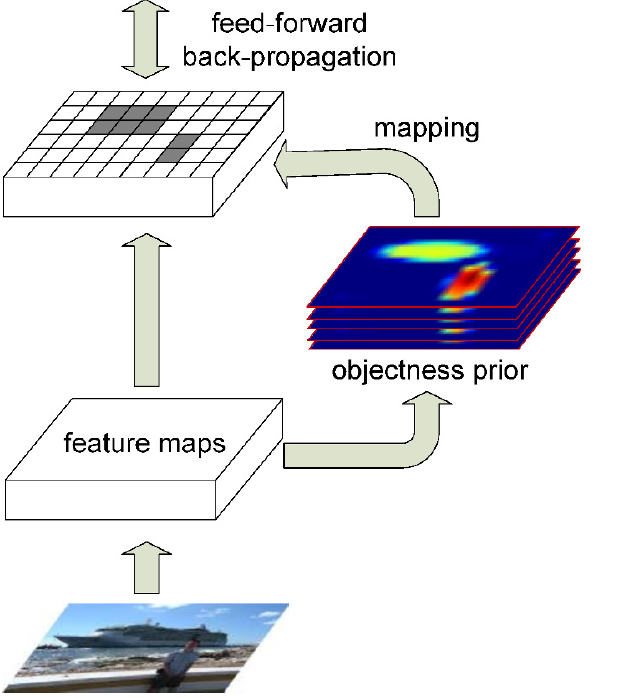}
\end{center}
\vskip -0.1 in
   \caption{Mapping the objectness prior with object detection. We first binarize the objectness prior maps according to $o_p$, then project the binary masks to the detection domain of the last convolutional feature maps. The locations within the masks are collected for detecting objects.}
\label{mapping}
\vskip -0.2 in
\label{mapppp}
\end{figure}

By now, each box has its objectness label and class-specific label. The network will update the class-specific labels dynamically for assisting object detection with objectness prior at training phase. For each mini-batch at feed-forward time, the network runs both the objectness prior and class-specific detection. But at the back-propagation phase, the network firstly generates the objectness prior, then for detection, samples whose objectness scores are high than threshold $o_p$ are selected (Figure \ref{mapppp}). The extra computation only comes from the selection of samples for back-propagation. With suitable $o_p$ (we use $o_p$ = 0.03 for all models), only a small number of samples is selected for updating the detection branch, thus the complexity of backward pass has been significantly reduced.

\section{Training and Testing}

In this section, we firstly introduce the multi-task loss function for optimizing the networks. Then we explain how to optimize the network jointly and perform inference directly.

\subsection{Loss Function}

For each location, our network has three sibling output branches. The first outputs the objectness confidence score $p^{obj} = \{p_{0}^{obj}, p_{1}^{obj}\}$, computed by a Softmax over the 2$\times A$ outputs of objectness prior ($A$ = 10 in this paper, as there are 10 types of default boxes). We denote the objectness loss with $L_{obj}$. The second branch outputs the bounding-box regression loss, denoted by $L_{loc}$. It targets at minimizing the smoothed $L_1$ loss \cite{frcnn} between the predicted location offsets $t = (t_{x},t_{y},t_{w},t_{h})$ and the target offsets $t^* = (t^*_{x},t^*_{y},t^*_{w},t^*_{h})$. Different from Fast R-CNN \cite{frcnn} that regresses the offsets for each of $K$ classes, we just regress the location one time with no class-specific information. The third branch outputs the classification loss $L_{cls|obj}$ for each box,  over $K$$+1$ categories. Given the objectness confidence score $p^{obj}$, the branch first excludes regions whose scores are lower than the threshold $o_{p}$.  Then like $L_{obj}$,  $L_{cls|obj}$ is computed by a Softmax over $K$$+1$ outputs for each location $p^{cls|obj} = \{p_{0}^{cls|obj}, p_{1}^{cls|obj}, \dots, p_{K}^{cls|obj}\}$. We use a multi-task loss $L$ to jointly train the networks end-to-end for objectness prior, classification and bounding-box regression:
\begin{equation}
L = \alpha \frac{1}{N_{obj}}L_{obj} + \beta  \frac{1}{N_{loc}} L_{loc} + (1-\alpha-\beta) \frac{1}{N_{cls|obj}} L_{cls|obj}.
\label{loss}
\end{equation}
The hyper-parameters $\alpha$ and $\beta$ in Equation \ref{loss} control the balance between the three losses. We normalize each loss term with its input number. Under this normalization, $\alpha = \beta = \frac{1}{3}$ works well and is used in all experiments.

\subsection{Joint Training and Testing}

\begin{table*}[t]\scriptsize\centering
\begin{center}
%p{1.5cm}|p{0.3cm}p{0.3cm}p{0.3cm}p{0.3cm}p{0.3cm}p{0.3cm}p{0.25cm}p{0.25cm}p{0.4cm}p{0.3cm}p{0.35cm}p{0.3cm}p{0.45cm}p{0.4cm}p{0.4cm}p{0.4cm}p{0.4cm}p{0.3cm}p{0.3cm}p{0.3cm}|p{0.4cm}
\begin{spacing}{1.25}
\begin{tabular}{p{1.85cm}|p{0.43cm}|p{0.25cm}p{0.25cm}p{0.25cm}p{0.25cm}p{0.25cm}p{0.25cm}p{0.25cm}p{0.25cm}p{0.25cm}p{0.25cm}p{0.25cm}p{0.25cm}p{0.25cm}p{0.3cm}p{0.3cm}p{0.3cm}p{0.25cm}p{0.25cm}p{0.25cm}c}
\scriptsize Method&\scriptsize\centering mAP&\centering\scriptsize aero&\centering\scriptsize bike&\centering\scriptsize bird&\centering\scriptsize boat&\centering\scriptsize bottle&\centering\scriptsize bus&\centering\scriptsize car&\centering\scriptsize cat& \centering\scriptsize chair&\centering\scriptsize cow&\centering\scriptsize table&\centering\scriptsize dog&\centering\scriptsize horse&\centering\scriptsize mbike&\centering\scriptsize person&\centering\scriptsize plant&\centering\scriptsize sheep&\centering\scriptsize sofa&\centering\scriptsize train&\scriptsize tv \\
\hline
  Fast R-CNN\cite{frcnn} &70.0& 77.0 & 78.1& 69.3& 59.4& 38.3& 81.6 & 78.6& 86.7 & 42.8& 78.8& 68.9 & 84.7& 82.0& 76.6& 69.9& 31.8& 70.1& 74.8& 80.4 & 70.4 \\
  Faster R-CNN\cite{fasterrcnn}&73.2& 76.5& 79.0& 70.9& 65.5& 52.1& 83.1& 84.7& 86.4& 52.0& \textbf{81.9}& 65.7& 84.8& 84.6& 77.5& 76.7& 38.8& 73.6& 73.9& 83.0& 72.6 \\
  SSD300\cite{ssd}&      72.1& 75.2& 79.8& 70.5& 62.5& 41.3& 81.1& 80.8& 86.4& 51.5& 74.3& 72.3& 83.5& 84.6& 80.6& 74.5& 46.0& 71.4& 73.8& 83.0& 69.1 \\
  SSD500\cite{ssd}&      75.1& 79.8& 79.5& 74.5& 63.4& 51.9& 84.9& \textbf{85.6}& 87.2& 56.6& 80.1& 70.0& 85.4& 84.9& 80.9& 78.2& 49.0& \textbf{78.4}& 72.4& 84.6& 75.5 \\
  \hline
  RON320&    74.2& 75.7& 79.4& 74.8& 66.1& 53.2& 83.7& 83.6& 85.8& 55.8& 79.5& 69.5& 84.5& 81.7& 83.1& 76.1& 49.2& 73.8& 75.2& 80.3& 72.5 \\
  RON384&    75.4& 78.0& 82.4& 76.7& 67.1& 56.9& 85.3& 84.3& 86.1& 55.5& 80.6& 71.4& 84.7& 84.8& 82.4& 76.2& 47.9& 75.3& 74.1& 83.8& 74.5 \\
  RON320++&    76.6& 79.4& \textbf{84.3}& 75.5& \textbf{69.5}& 56.9& 83.7& 84.0& \textbf{87.4}& 57.9& 81.3& \textbf{74.1}& 84.1& 85.3& \textbf{83.5}& 77.8& 49.2& 76.7& \textbf{77.3}& \textbf{86.7}& 77.2 \\
  RON384++&    \textbf{77.6}& \textbf{86.0}& 82.5& \textbf{76.9}& 69.1& \textbf{59.2}& \textbf{86.2}& 85.5& 87.2& \textbf{59.9}& 81.4& 73.3& \textbf{85.9}& \textbf{86.8}& 82.2& \textbf{79.6}& \textbf{52.4}& 78.2& 76.0& 86.2& \textbf{78.0} \\

\end{tabular}
\end{spacing}
\end{center}
\vskip -0.2 in
\caption{Detection results on PASCAL VOC 2007 test set. The entries with the best APs for each object category are bold-faced.}
\label{voc07}
\end{table*}

\begin{figure*}[htb]
\centering
\begin{minipage}[t]{0.3\linewidth}
\centering
\includegraphics[width=0.9\linewidth]{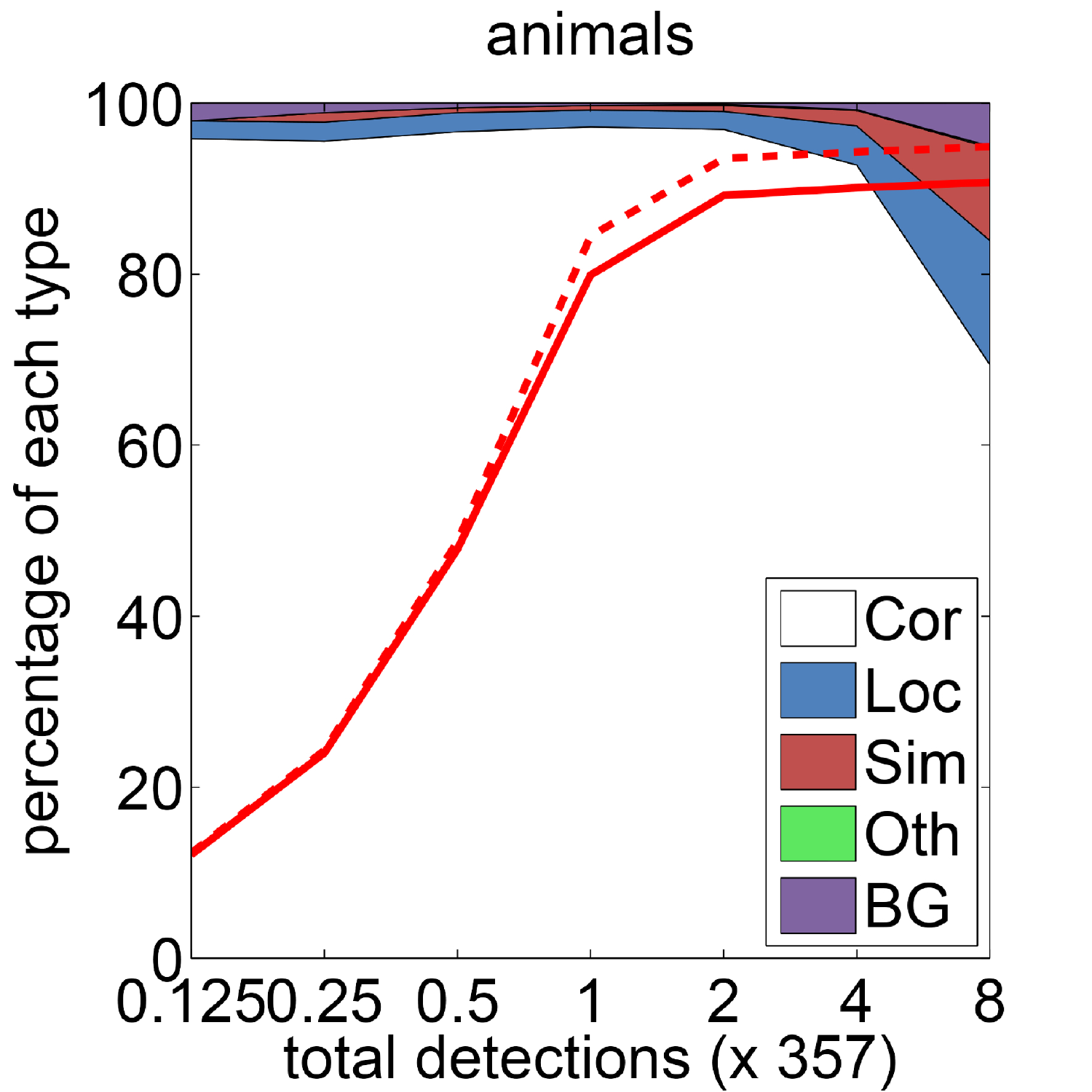}
\end{minipage}
\begin{minipage}[t]{0.3\linewidth}
\centering
\includegraphics[width=0.9\linewidth]{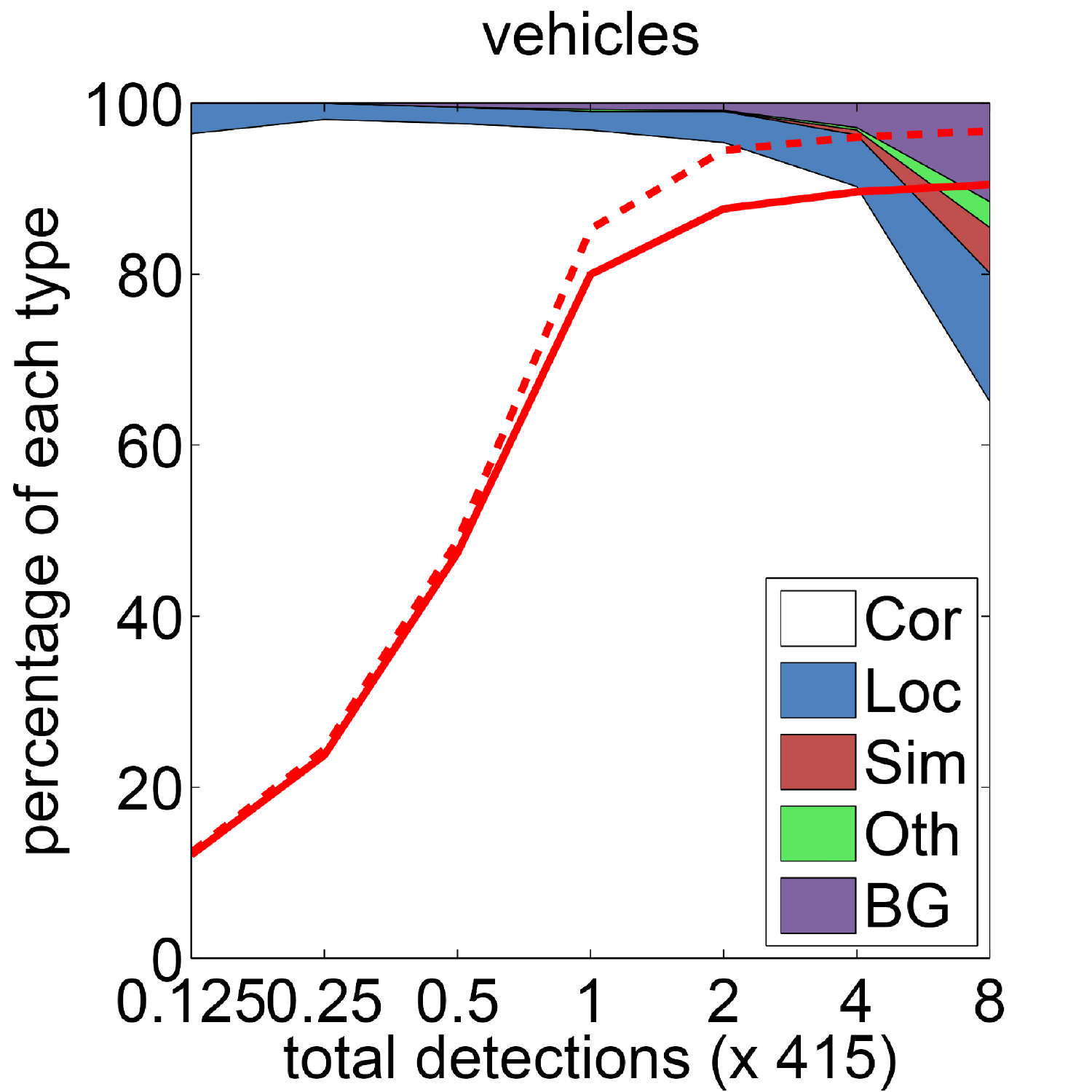}
\end{minipage}
\begin{minipage}[t]{0.3\linewidth}
\centering
\includegraphics[width=0.9\linewidth]{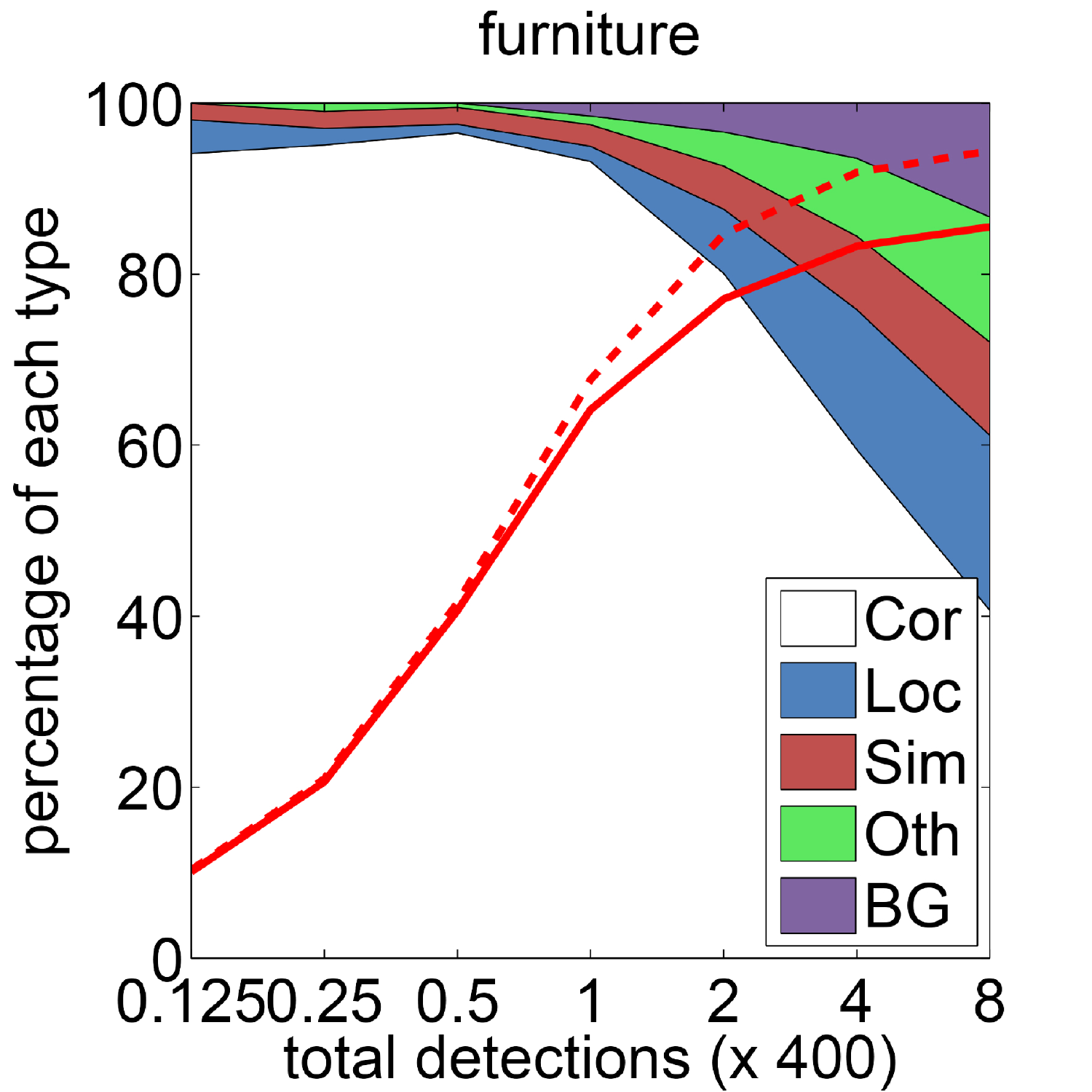}
\end{minipage}
\caption{Visualization of performance for RON384 on animals, vehicles, and furniture from VOC2007 test. The Figures show the cumulative fraction of detections that are correct (Cor) or false positive due to poor localization (Loc), confusion with similar categories (Sim), with others (Oth), or with background (BG). The solid red line reflects the change of recall with the `strong' criteria (0.5 jaccard overlap) as the number of detections increases. The dashed red line uses the `weak' criteria (0.1 jaccard overlap).}
\label{det_error}
\vskip -0.2 in
\end{figure*}

We optimize the models described above jointly with end-to-end training.  After the matching step, most of the default boxes are negatives, especially when the number of default boxes is large. Here, we introduce a dynamic training strategy to scan the negative space. At training phase, each SGD mini-batch is constructed from $N$ images chosen uniformly from the dataset. At each mini-batch, (a) for objectness prior, all of the positive samples are selected for training. Negative samples are randomly selected from regions with negative labels such that the ratio between positive and negative samples is 1:3; (b) for detection, we first reduce the sample number according to objectness prior scores generated at this mini-batch (as discribed in Section \ref{combine}). Then all of the positive samples are selected. We randomly select negative samples such that the ratio between positive and negative samples is 1:3. We note that recent works such as Faster R-CNN \cite{fasterrcnn} and R-FCN \cite{rfcn} usually use multi-stage training for joint optimization. In contrast, the loss function \ref{loss} is trained end-to-end in our implementation by back-propagation and SGD, which is more efficient at training phase. At the beginning of training, the objectness prior maps are in chaos. However, along with the training progress, the objectness prior maps are more concentrated on areas covering objects.

\textbf{Data augmentation}
To make the model more robust to various object scales, each training image is randomly sampled by one of the following options: (i) Using the original/flipped input image; (ii) Randomly sampling a patch whose edge length is \{$\frac{4}{10}$, $\frac{5}{10}$, $\frac{6}{10}$, $\frac{7}{10}$, $\frac{8}{10}$, $\frac{9}{10}$\} of the original image and making sure that at least one object's center is within this patch. We note that the data augmentation strategy described above will increase the number of large objects, but the optimization benefit for small objects is limited. We overcome this problem by adding a small scale for training. A large object at one scale will be smaller at smaller scales. This training strategy could effectively avoid over-fitting of objects with specific sizes.

\textbf{Inference}
At inference phase, we multiply the class-conditional probability and the individual box confidence predictions. The class-specific confidence score for each box is defined as Equation \ref{det_score}:
\begin{equation}
p^{cls} = p^{obj} \cdot p^{cls|obj}.
\label{det_score}
\end{equation}
The scores encode both the probability of the class appearing in the box and how well the predicted box fits the object. After the final score of each box is generated, we adjust the boxes according to the bounding box regression outputs. Finally, non-maximum suppression is applied to get the final detection results.

\section{Results}

We train and evaluate our models on three major datasets: PASCAL VOC 2007, PASCAL VOC 2012, and MS COCO. For fair comparison, all experiments are based on the VGG-16 networks. We train all our models on a single Nvidia TitanX GPU, and demonstrate state-of-the-art results on all three datasets.

\subsection{PASCAL VOC 2007}

\begin{table*}[t]\scriptsize \centering
\begin{center}
%p{1.5cm}|p{0.3cm}p{0.3cm}p{0.3cm}p{0.3cm}p{0.3cm}p{0.3cm}p{0.25cm}p{0.25cm}p{0.4cm}p{0.3cm}p{0.35cm}p{0.3cm}p{0.45cm}p{0.4cm}p{0.4cm}p{0.4cm}p{0.4cm}p{0.3cm}p{0.3cm}p{0.3cm}|p{0.4cm}
\begin{spacing}{1.25}
\begin{tabular}{p{1.9cm}|p{0.43cm}|p{0.25cm}p{0.25cm}p{0.25cm}p{0.25cm}p{0.25cm}p{0.25cm}p{0.25cm}p{0.25cm}p{0.25cm}p{0.25cm}p{0.25cm}p{0.25cm}p{0.25cm}p{0.3cm}p{0.3cm}p{0.3cm}p{0.25cm}p{0.25cm}p{0.25cm}c}

 \scriptsize Method&\scriptsize\centering mAP&\centering\scriptsize aero&\centering\scriptsize bike&\centering\scriptsize bird&\centering\scriptsize boat&\centering\scriptsize bottle&\centering\scriptsize bus&\centering\scriptsize car&\centering\scriptsize cat& \centering\scriptsize chair&\centering\scriptsize cow&\centering\scriptsize table&\centering\scriptsize dog&\centering\scriptsize horse&\centering\scriptsize mbike&\centering\scriptsize person&\centering\scriptsize plant&\centering\scriptsize sheep&\centering\scriptsize sofa&\centering\scriptsize train&\scriptsize tv \\
\hline
  Fast R-CNN\cite{frcnn}   &68.4&	82.3& 78.4& 70.8& 52.3& 38.7& 77.8& 71.6& 89.3& 44.2& 73.0& 55.0& 87.5&	80.5& 80.8&	72.0& 35.1&	68.3& 65.7&	80.4& 64.2 \\
  OHEM\cite{ohem} &71.9& 83.0& 81.3& 72.5& 55.6& 49.0& 78.9& 74.7& 89.5& 52.3& 75.0& 61.0& \textbf{87.9}& 80.9& 82.4& 76.3& 47.1& 72.5& 67.3& 80.6& \textbf{71.2} \\
  Faster R-CNN\cite{fasterrcnn} &70.4&	84.9& 79.8&	74.3& 53.9&	49.8& 77.5&	75.9& 88.5&	45.6& 77.1&	55.3& 86.9&	81.7& 80.9&	79.6& 40.1&	72.6& 60.9&	81.2& 61.5 \\
  HyperNet\cite{hypernet} & 	71.4	&84.2	&78.5	&73.6&	55.6	&53.7&	78.7	&79.8	&87.7&	49.6&	74.9	&52.1	&86.0&	81.7	& 83.3	&\textbf{81.8}	&48.6	&73.5&	59.4	&79.9	&65.7\\	
  SSD300\cite{ssd}       &70.3&	84.2& 76.3&	69.6& 53.2&	40.8& 78.5&	73.6& 88.0&	50.5& 73.5&	61.7& 85.8&	80.6& 81.2&	77.5& 44.3&	73.2& 66.7&	81.1& 65.8 \\
  SSD500\cite{ssd}       &73.1&	84.9& 82.6&	74.4& 55.8&	50.0& 80.3&	78.9& 88.8&	53.7& 76.8&	59.4& 87.6&	83.7& 82.6&	81.4& 47.2&	75.5& 65.6&	84.3& 68.1 \\
\hline
  RON320    &71.7&	84.1& 78.1& 71.0& 56.8& 46.9& 79.0& 74.7& 87.5& 52.5& 75.9& 60.2& 84.8&	79.9& 82.9&	78.6& 47.0&	75.7& 66.9&	82.6& 68.4 \\
  RON384    &73.0&	85.4& 80.6& 71.9& 56.3& 49.8& 80.6& 76.8& 88.2& 53.6& 78.1& 60.4& 86.4&	81.5& 83.8&	79.4& 48.6&	77.4& 67.7&	83.4& 69.5 \\
  RON320++  &74.5&	\textbf{87.1}& 81.0& 74.6& 58.8& 51.7& \textbf{82.1}& 77.0& 89.7& 57.2& 79.9& \textbf{62.6}& 87.2&	83.2& \textbf{85.0}&	80.5& 51.4&	76.7& 68.5&	\textbf{84.8}& 70.4 \\
  RON384++  &\textbf{75.4}&	86.5& \textbf{82.9}& \textbf{76.6}& \textbf{60.9}& \textbf{55.8}& 81.7& \textbf{80.2}& \textbf{91.1}& \textbf{57.3}& \textbf{81.1}& 60.4& 87.2&	\textbf{84.8}& 84.9&	81.7& \textbf{51.9}&	\textbf{79.1}& \textbf{68.6}&	84.1& 70.3 \\
\end{tabular}
\end{spacing}
\end{center}
\vskip -0.2 in
\caption{Results on PASCAL VOC 2012 test set. All methods are based on the pre-trained VGG-16 networks.}
\label{voc12}
\vskip -0.1 in
\end{table*}

On this dataset, we compare RON against seminal Fast R-CNN \cite{frcnn}, Faster R-CNN \cite{fasterrcnn}, and the most recently proposed SSD \cite{ssd}. All methods are trained on VOC2007 trainval and VOC2012 trainval, and tested on VOC2007 test dataset. During training phase, we initialize the parameters for all the newly added layers by drawing weights from a zero-mean Gaussian distribution with standard deviation 0.01. All other layers are initialized by standard VGG-16 model \cite{frcnn}. We use the $10^{-3}$ learning rate for the first 90k iterations, then we decay it to $10^{-4}$ and continue training for next 30k iterations. The batch size is 18 for $320$$\times320$ model according to the GPU capacity. We use a momentum of 0.9 and a weight decay of 0.0005.

Table \ref{voc07} shows the result comparisons of the methods\footnote{We note that the latest SSD uses new training tricks (color distortion, random expansion and online hard example mining), which makes the results much better. We expect these tricks will also improve our results, which is beyond the focus of this paper.}. With 320$\times$320 input size, RON is already better than Faster R-CNN. By increasing the input size to 384$\times$384, RON gets 75.4\% mAP, outperforming Faster R-CNN by a margin of 2.2\%. RON384 is also better than SSD with input size 500$\times$500. Finally, RON could achieve high mAPs of 76.6\% (RON320++) and 77.6\% (RON384++) with multi-scale testing, bounding box voting and flipping \cite{ion}.

Small objects are challenging for detectors. As shown in Table \ref{voc07}, all methods have inferior performance on `boat' and `bottle'. However, RON improves performance of these categories by significant margins: 4.0 points improvement for `boat' and 7.1 points improvement for `bottle'. In summary, performance of 17 out of 20 categories has been improved by RON.

To understand the performance of RON in more detail, we use the detection analysis tool from \cite{detann}. Figure \ref{det_error} shows that our model can detect various object categories with high quality. The recall is higher than 85\%, and is much higher with the `weak' (0.1 jaccard overlap) criteria.

\subsection{PASCAL VOC 2012}

We compare RON against top methods on the comp4 (outside data) track from the public leaderboard on PASCAL VOC 2012. The training data is the union set of all VOC 2007, VOC 2012 train and validation datasets, following \cite{fasterrcnn}\cite{frcnn}\cite{ssd}. We see the same performance trend as we observed on VOC 2007 test. The results, as shown in Table \ref{voc12}, demonstrate that our model performs the best on this dataset. Compared with Faster R-CNN and other variants \cite{ohem}\cite{hypernet}, the proposed network is significantly better, mainly due to the reverse connection and the use of boxes from multiple feature maps.

\subsection{MS COCO}

To further validate the proposed framework on a larger and more challenging dataset, we conduct experiments on MS COCO \cite{coco} and report results from test-dev2015 evaluation server. The evaluation metric of MS COCO dataset is different from PASCAL VOC. The average mAP over different IoU thresholds, from 0.5 to 0.95 (written as "0.5:0.95") is the overall performance of methods. This places a significantly larger emphasis on localization compared to the PASCAL VOC metric which only requires IoU of 0.5. We use the 80k training images  and 40k validation images \cite{fasterrcnn}  to train our model, and validate the performance on the test-dev2015 dataset which contains 20k images.
We use the $5$$\times10^{-4}$ learning rate for 400k iterations, then we decay it to $5$$\times 10^{-5}$ and continue training for another 150k iterations. As instances in MS COCO dataset are smaller and denser than those in PASCAL VOC dataset, the minimum scale $s_{min}$ of the referenced box size is 24 for 320$\times$320 model, and 32 for 384$\times$384 model. Other settings are the same as PASCAL VOC dataset.
\begin{table}[h]\small
\begin{center}
\begin{tabular}{l|c|ccc}
\multirow{2}{*}{Method}&\multirow{2}{*}{Train Data}&\multicolumn{3}{c}{Average Precision} \\
 & & 0.5 & 0.75 &0.5:0.95 \\
\hline
Fast R-CNN\cite{frcnn}&train&35.9& - & 19.7\\
OHEM\cite{ohem}&trainval&42.5& 22.2 & 22.6\\
OHEM++\cite{ohem}&trainval&45.9& 26.1 & 25.5\\
Faster R-CNN\cite{fasterrcnn}&trainval&42.7& - & 21.9\\
SSD300\cite{ssd}&trainval35k&38.0& 20.5 & 20.8\\
SSD500\cite{ssd}&trainval35k&43.7& 24.7 & 24.4\\
\hline
RON320&trainval&44.7& 22.7 & 23.6\\
RON384&trainval&46.5& 25.0 & 25.4\\
RON320++&trainval&47.5& 25.9 & 26.2\\
RON384++&trainval&\textbf{49.5}& \textbf{27.1} & \textbf{27.4}\\
\end{tabular}
\end{center}
\vskip -0.1 in
\caption{MS COCO test-dev2015 detection results.}
\label{time_comp}
\vskip -0.2 in
\end{table}

With the standard COCO evaluation metric, Faster R-CNN scores 21.9\% AP, and RON improves it to 27.4\% AP. Using the VOC overlap metric of IoU $\geq$0.5, RON384++ gives a 5.8 points boost compared with SSD500. It is also interesting to note that with 320$\times$320 input size, RON gets 26.2\% AP, improving the SSD with 500$\times$500 input size by 1.8 points on the strict COCO AP evaluation metric.

We also compare our method against Fast R-CNN with online hard example mining (OHEM) \cite{ohem}, which gives a considerable improvement on Fast R-CNN. The OHEM method also adopts recent bells and whistles to further improve the detection performance. The best result of OHEM is 25.5\% AP (OHEM++). RON gets 27.4\% AP, which demonstrates that the proposed network is more competitive on large dataset.

\subsection{From MS COCO to PASCAL VOC}

Large-scale dataset is important for improving deep neural networks. In this experiment, we investigate how the MS COCO dataset can help with the detection performance of PASCAL VOC. As the categories on MS COCO are a superset of these on PASCAL VOC dataset, the fine-tuning process becomes easier compared with the ImageNet pre-trained model. Starting from MS COCO pre-trained model, RON leads to 81.3\% mAP on PASCAL VOC 2007 and 80.7\% mAP on PASCAL VOC 2012.

\begin{table}[h]
\begin{center}
\begin{tabular}{p{2.8cm}|p{2cm}<{\centering}|p{2cm}<{\centering}}
 Method  & 2007 test & 2012 test\\
\hline
Faster R-CNN\cite{fasterrcnn}&78.8 &75.9 \\
OHEM++\cite{ohem}&- &80.1 \\
SSD512\cite{ssd}&- &80.0 \\
\hline
RON320&78.7 &76.3 \\
RON384&80.2 &79.0 \\
RON320++&80.3 &78.7 \\
RON384++&\textbf{81.3} &\textbf{80.7} \\
\end{tabular}
\end{center}
\vskip -0.05 in
\caption{The performance on PASCAL VOC datasets. All models are pre-trained on MS COCO, and fine-tuned on PASCAL VOC.}
\label{coco_voc}
\vskip -0.2 in
\end{table}

The extra data from the MS COCO dataset increases the mAP by 3.7\% and 5.3\%. Table \ref{coco_voc} shows that the model trained on COCO+VOC has
the best mAP on PASCAL VOC 2007 and PASCAL VOC 2012. When submitting, our model with 384$\times$384 input size has been ranked as the top 1 on the VOC 2012 leaderboard among VGG-16 based models. We note that other public methods with better results are all based on much deeper networks \cite{resnet}.

\section{Ablation Analysis}

\subsection{Do Multiple Layers Help?}

As described in Section \ref{network}, our networks generate detection boxes from multiple layers and combine the results. In this experiment, we compare how layer combinations affect the final performance. For all of the following experiments as shown in Table \ref{layercombine}, we use exactly the same settings and input size (320$\times$320), except for the layers for object detection.

\begin{table}[h]\small
\begin{center}
\begin{tabular}{cccc|c}
\hline
\multicolumn{4}{c|}{detection from layer } &\multirow{2}{*}{mAP}\\
  4&5 & 6 &  7 \\
\hline
& &  &   $\checkmark$  &65.6 \\
& &$\checkmark$&$\checkmark$ & 68.3 \\
&$\checkmark$&$\checkmark$&$\checkmark$ & 72.5 \\
$\checkmark$&$\checkmark$&$\checkmark$&$\checkmark$& 74.2 \\
\hline
\end{tabular}
\end{center}
\vskip -0.05 in
\caption{Combining features from different layers.}
\label{layercombine}
\vskip -0.1 in
\end{table}

From Table \ref{layercombine}, we see that it is necessary to use all of the layer 4, 5, 6 and 7 such that the detector could get the best performance.

\subsection{Objectness Prior}

\begin{figure}[h]
\begin{center}
    \includegraphics[width=0.92\linewidth]{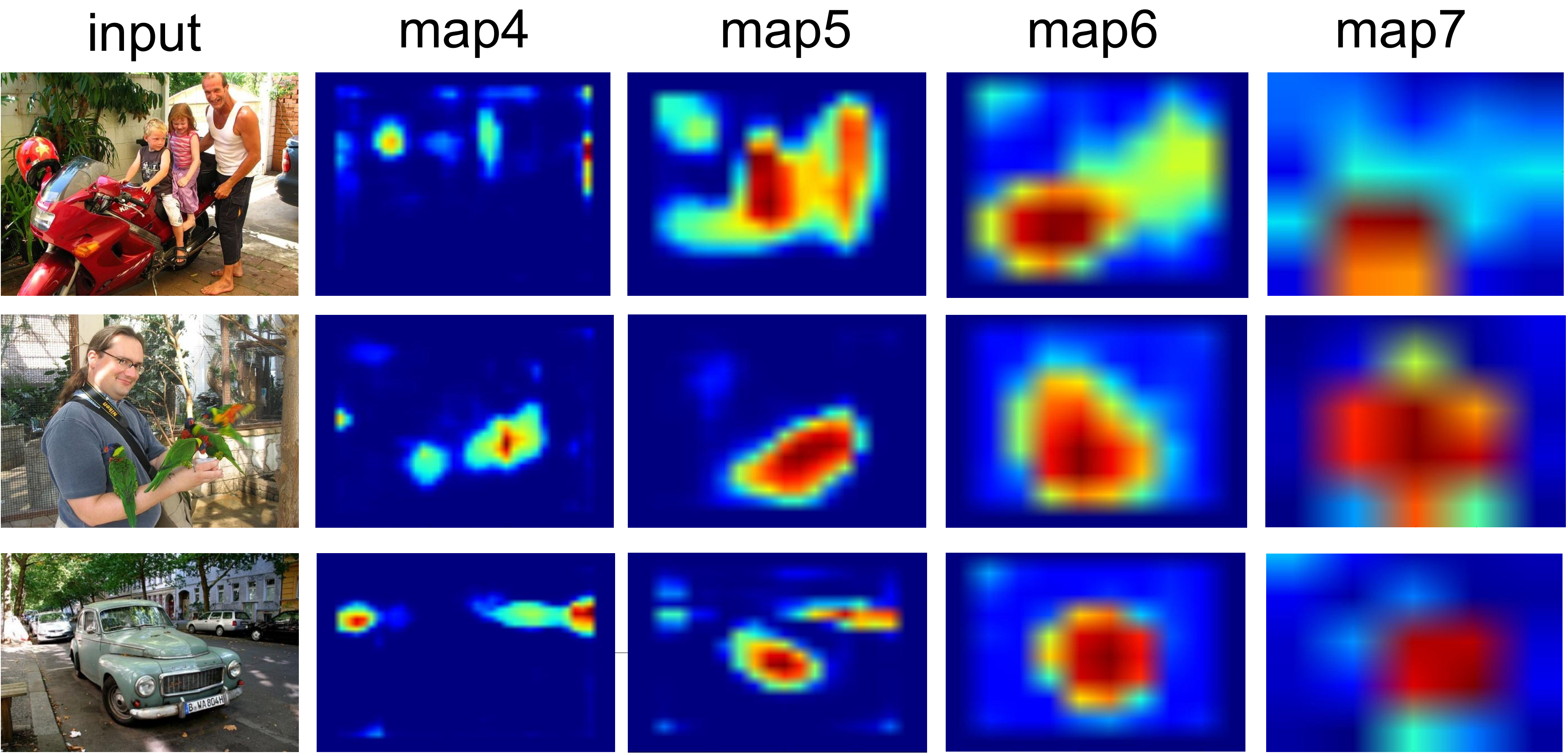}
\end{center}
\vskip -0.1 in
   \caption{Objectness prior maps generated from images.}
\label{objectness}
\vskip -0.1 in
\end{figure}

As introduced in Section \ref{objprior}, the network generates objectness prior for post detection. The objectness prior maps involve not only the strength of the responses, but also their spatial positions. As shown in Figure \ref{objectness}, objects with various scales will respond at the corresponding maps. The maps can guide the search of different scales of objects, thus significantly reducing the searching space.

We also design an experiment to verify the effect of objectness prior. In this experiment, we remove the objectness prior module and predict the detection results only from the detection module. Other settings are exactly the same as the baseline. Removing objectness prior maps leads to 69.6\% mAP on VOC 2007 test dataset, resulting 4.6 points drop from the 74.2\% mAP baseline.

\subsection{Generating Region Proposals}

\begin{figure}[h]
\begin{center}
    \includegraphics[width=0.8\linewidth]{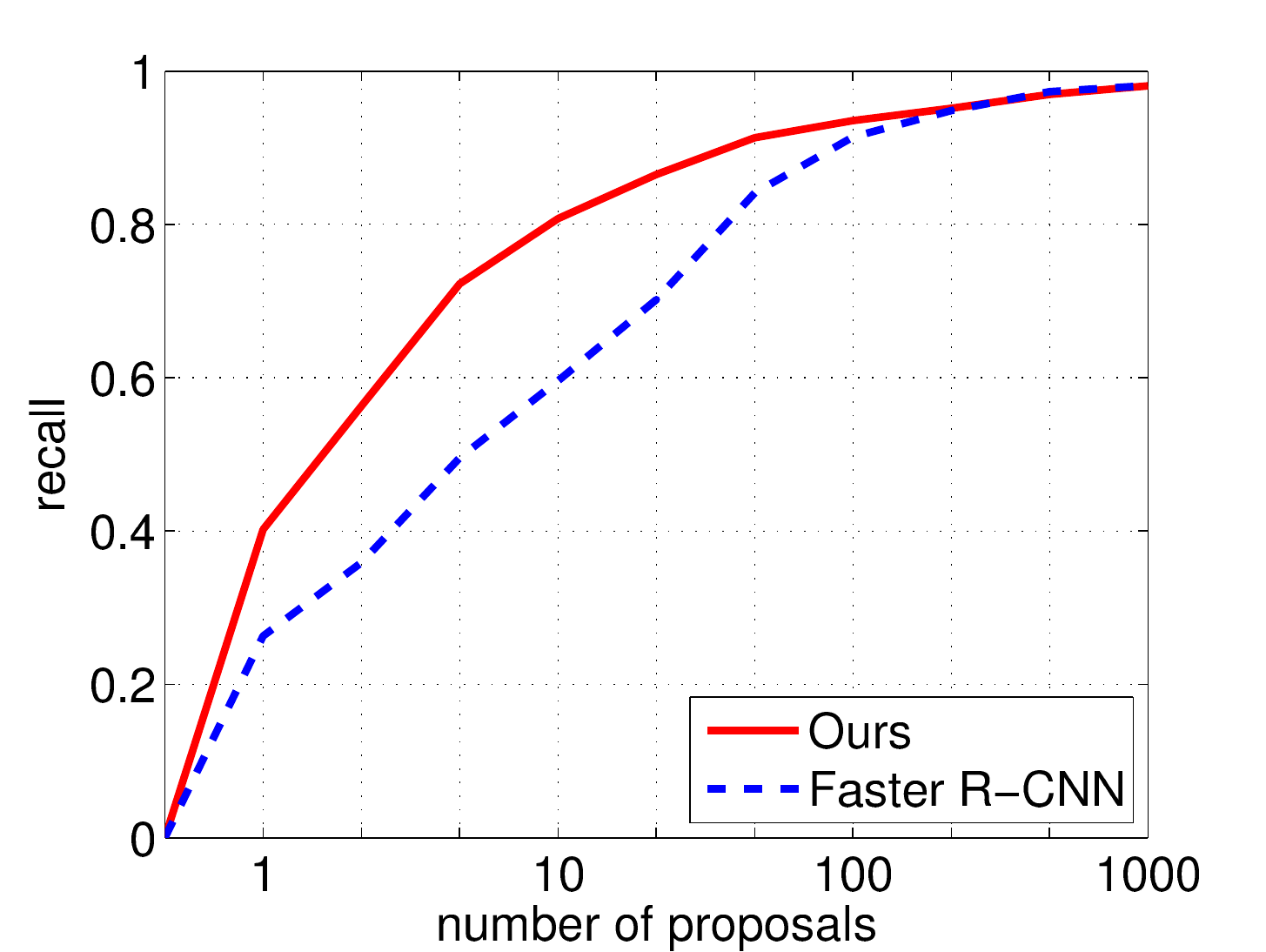}
\end{center}
\vskip -0.1 in
   \caption{Recall versus number of proposals on the PASCAL VOC 2007 test set (with IoU = 0.5).}
\label{recall}
\vskip -0.1 in
\end{figure}

After removing the detection module, our network could get region proposals. We compare the proposal performance against Faster R-CNN \cite{fasterrcnn} and evaluate recalls with different numbers of proposals on PASCAL VOC 2007 test set, as shown in Figure \ref{recall}. The $N$ proposals are the top-$N$ ranked ones based on the confidence generated by these methods.

Both Faster R-CNN and RON achieve promising region proposals when the region number is larger than 100. However, with fewer region proposals, the recall of RON boosts Faster R-CNN  by a large margin. Specifically, with top 10 region proposals, our 320 model gets 80.7\% recall, outperforming Faster R-CNN by 20 points. This validates that our model is more effective in applications with less region proposals.

\section{Conclusion}

We have presented RON, an efficient and effective object detection framework. We design the reverse connection to enable the network to detect objects on multi-levels of CNNs. And the objectness prior is also proposed to guide the search of objects. We optimize the whole networks by a multi-task loss function, thus the networks can directly predict final detection results. On standard benchmarks, RON achieves state-of-the-art object detection performance.

\begin{figure*}
\begin{center}
    \includegraphics[width=0.9\linewidth]{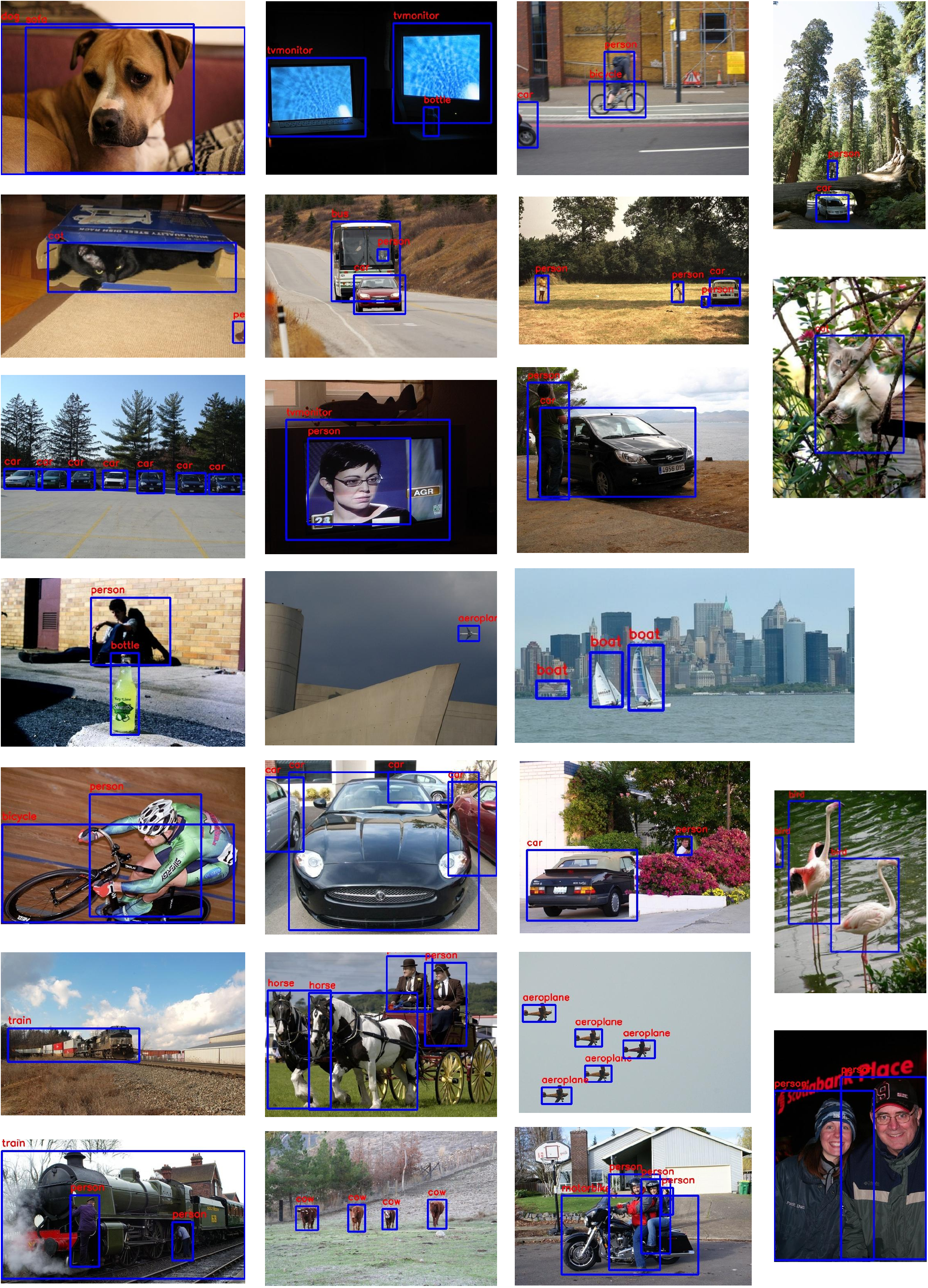}
\end{center}
   \caption{Selected examples of object detection results on the PASCAL VOC 2007 test set using the RON320 model. The training data is 07+12 trainval (74.2\% mAP on the 2007 test set).}
\label{det_example}
\end{figure*}

{\small
\bibliographystyle{ieee}
\bibliography{egbib}
}

\end{document}